\documentclass[10pt]{article}
\usepackage[accepted]{tmlr}

\makeatletter
\def\@startauthor{\noindent\normalsize\bf\ignorespaces}
\makeatother

\usepackage{amsmath,amsfonts,bm}

\def\eqref#1{equation~\ref{#1}}

\def\1{\bm{1}}

\DeclareMathAlphabet{\mathsfit}{\encodingdefault}{\sfdefault}{m}{sl}
\SetMathAlphabet{\mathsfit}{bold}{\encodingdefault}{\sfdefault}{bx}{n}

\usepackage{hyperref}
\hypersetup{hidelinks}
\usepackage{url}
\usepackage{booktabs}
\usepackage{multirow}
\usepackage{algorithm}

\usepackage{algorithmic}
\usepackage{subcaption}
\usepackage{graphicx}
\usepackage{amssymb}

\newcommand{\sd}{\sigma_d}
\newcommand{\Fth}{F_\theta}
\newcommand{\Fthm}{F_{\theta^-}}
\newcommand{\fth}{f_\theta}
\newcommand{\fthm}{f_{\theta^-}}
\newcommand{\cnoise}{c_{\text{noise}}}
\newcommand{\cskip}{c_{\text{skip}}}
\newcommand{\cout}{c_{\text{out}}}
\newcommand{\cin}{c_{\text{in}}}
\newcommand{\dt}{\Delta t}
\newcommand{\tk}{t_k}
\newcommand{\tmax}{t_{\max}}
\newcommand{\pnorm}{\mathrm{pnorm}}
\newcommand{\norm}{\mathrm{norm}}

\title{Stable Continuous-Time Consistency Distillation:\\
An Empirical Study with a Multistep Extension}

\author{
\name Vinay Saji Mathew \email vvm5242@psu.edu \\
\addr The Pennsylvania State University \\[0.6em]
\rm\name Soundar R.\ Kumara \email skumara@psu.edu \\
\addr The Pennsylvania State University \\[0.6em]
\rm\name Gretta D.\ Kellogg \email gretta.kellogg@psu.edu \\
\addr The Pennsylvania State University \\[0.6em]
\rm\name William KM Lai
\thanks{Part of the work was done while William KM Lai was at Cornell University.
Correspondence to: wklai2@buffalo.edu} \email wklai2@buffalo.edu \\
\addr State University of New York at Buffalo
}

\def\month{09}
\def\year{2026}
\def\openreview{\url{https://openreview.net/forum?id=di6ofoWEU8}}

\begin{document}

\maketitle

\begin{abstract}
  Continuous-time consistency models are an attractive route to few-step generation. They remove the discretization error inherent to the timestep grid that discrete-time consistency models depend on. Until recently, however, training them was unstable. \citet{lu2025simplifying} address this with the TrigFlow formulation and a coordinated set of architectural and training-objective changes. We conduct an independent empirical study of this framework on CIFAR-10 and ImageNet-64. No open-weight TrigFlow teachers exist at these scales, so we train them from scratch and then verify the consistency distillation (sCD) and consistency training (sCT) behavior of students initialized from these teachers. On CIFAR-10, our students reproduce the reported FID at one and two steps. On ImageNet-64, they fall short, a gap that appears related to our teacher and to architectural choices the paper leaves unspecified. We also report a parameter-count gap of about $5\%$ between the model size stated in the paper and a faithful re-implementation, which we attribute to the scale-and-shift conditioning projections in Adaptive Double Normalization. We then propose \emph{Multistep sCD (MS-sCD)}, a continuous-time, segment-conditioned multistep generalization of sCD. The segment-conditioning idea is borrowed from the discrete-time multistep consistency distillation of \citet{heek2024multistep}; we build on their preliminary continuous-time experiments within the TrigFlow/sCM framework. The rest of the algorithm follows sCD. MS-sCD replaces sCM's two-step intermediate time with a fixed segment-boundary schedule and recovers sCD at $M=1$. We report FID at $M=2,4,8$ and compare against the discrete-time MSCD baseline. Code, model weights, and training scripts are available on \href{https://github.com/EpiGenomicsCode/edm2_sCM}{GitHub} \citep{mathew2026software}.
\end{abstract}

\section{Introduction}
\label{sec:intro}

\subsection{Motivation}
\label{sec:intro:motivation}

Diffusion models \citep{sohl2015deep,ho2020denoising,song2021score,karras2022elucidating} produce high-quality samples but require many sequential network evaluations. Distillation reduces this cost by training a student to generate samples in fewer steps, but methods differ in what they ask the student to learn. Progressive distillation \citep{salimans2022progressive} repeatedly halves the sampling steps by teaching each student to match two steps of the previous model. Consistency models \citep{song2023consistency} instead learn to map points on the same probability-flow ODE trajectory to a common endpoint, enabling one- or few-step generation with a single network. Later methods \citep{song2024improved,geng2024consistency} refine the loss, timestep sampling, and weighting, while multistep consistency distillation (MSCD) \citep{heek2024multistep} divides the trajectory into segments so that each prediction covers a shorter interval. MSCD uses DDIM \citep{song2021ddim} teacher steps within these segments and an aDDIM sampler to correct the variance shortfall of finite DDIM integration. A separate direction compares distributions rather than enforcing consistency along trajectories: variational score distillation \citep{wang2023prolificdreamer}, distribution matching distillation \citep{yin2024dmd}, and moment matching \citep{salimans2024multistep} use distribution-level objectives.

For consistency models, a remaining question is whether training can avoid the timestep grid and its discretization error. \citet{song2023consistency} propose a continuous-time formulation (their Appendix~B), but report training instability. The main iCT results of \citet{song2024improved} use discrete-time training. sCM \citep{lu2025simplifying} addresses this instability through the TrigFlow parameterization and coordinated changes to time conditioning, normalization, and the training objective, reporting stable continuous-time CMs at scales up to 1.5B parameters on ImageNet-512. We examine whether these results can be reproduced and whether segment-conditioned distillation extends the framework to larger sampling budgets.

\subsection{Contributions}
\label{sec:intro:contributions}
Our contributions are:
\begin{enumerate}
  \item \textbf{Reproduction with implementation clarifications.} We reproduce sCM's TrigFlow teacher and sCD/sCT students on CIFAR-10 (unconditional, DDPM++ backbone) and ImageNet-64 (class-conditional, EDM2-S backbone), following the two configurations sCM uses. We make explicit the implementation details required to run the continuous-time CM training loop.
  \item \textbf{Continuous-time MS-sCD (extension).} Building on \citet{heek2024multistep}'s preliminary experiments with segment-conditioned continuous-time training, we derive, implement, and evaluate this approach within the TrigFlow/sCM framework.
  \item \textbf{Comparisons with existing methods.} We compare MS-sCD with MSCD \citep{heek2024multistep} on the same EDM2-S backbone to assess their relative strengths at different sampling budgets. We also evaluate moment matching \citep{salimans2024multistep} on an ADM-style backbone to examine an alternative distillation objective.
\end{enumerate}

\section{Background}
\label{sec:background}

\subsection{Diffusion process and probability flow ODE}
\label{sec:background:diffusion}

Diffusion models \citep{sohl2015deep,ho2020denoising,song2021score} define a forward process that gradually corrupts a data sample $x_0$ with Gaussian noise according to a schedule $(\alpha_t, \sigma_t)$, producing $x_t = \alpha_t\,x_0 + \sigma_t\,z$ with $z\sim\mathcal{N}(0, I)$. \citet{song2021score} show that the associated marginals $p_t(x_t)$ are also generated by a deterministic probability flow ODE (PF-ODE), $dx_t/dt = f(t)\,x_t - \tfrac{1}{2}g(t)^2\,\nabla_{x_t}\log p_t(x_t)$, whose vector field is determined by the score. Sampling reduces to integrating this ODE from $t = T$ (approximately Gaussian) back to $t = 0$ (data). The EDM framework \citep{karras2022elucidating} adopts a variance-exploding schedule and parameterizes the denoiser as $D_\theta(x_t) = \cskip(t)\,x_t + \cout(t)\,F_\theta(\cin(t)\,x_t,\,\cnoise(t))$, with $F_\theta$ a neural network and $(\cskip, \cout, \cin, \cnoise)$ preconditioners chosen so that the training objective has unit variance across timesteps.

\subsection{Flow matching}
\label{sec:background:fm}

Flow matching \citep{lipman2022flow,liu2022flow,albergo2023stochastic} fixes an interpolation $x_t = \alpha_t\,x_0 + \sigma_t\,z$ between data and noise and trains a velocity field $v_\theta(x_t, t)$ to match $\dot{x}_t = \alpha_t'\,x_0 + \sigma_t'\,z$, so the PF-ODE becomes $dx_t/dt = v_\theta(x_t, t)$. Different choices of $(\alpha_t, \sigma_t)$ recover variance-preserving diffusion, rectified flow \citep{liu2022flow}, or stochastic interpolants \citep{albergo2023stochastic}.

\subsection{TrigFlow}
\label{sec:background:trigflow}

TrigFlow \citep{lu2025simplifying} unifies EDM and flow matching by parameterizing noisy samples as
\begin{equation}
  x_t = \cos(t)\,x_0 + \sin(t)\,z, \quad z \sim \mathcal{N}(0, \sd^2 I), \quad t \in [0, \pi/2].
  \label{eq:trigflow_process}
\end{equation}
The ideal velocity along this trajectory is $\dot{x}_t^* = \cos(t)\,z - \sin(t)\,x_0$. A TrigFlow network $\Fth$ approximates this velocity, trained via:
\begin{equation}
  \mathcal{L}_{\text{DM}} = \mathbb{E}_{x_0, z, t}\bigl[\|\sd\,\Fth(x_t/\sd,\, t) - \dot{x}_t^*\|^2\bigr].
  \label{eq:teacher_loss}
\end{equation}

The PF-ODE is $dx_t/dt = \sd\,\Fth(x_t/\sd, t)$. TrigFlow satisfies EDM's preconditioner requirements with $\cskip(t) = \cos(t)$, $\cout(t) = -\sd\sin(t)$, $\cin(t) = 1/\sd$, and $\cnoise(t) = t$ (identity time conditioning, see \citealt{lu2025simplifying} for the derivation).

\textbf{CM parameterization.} The TrigFlow CM \citep{lu2025simplifying} parameterizes the consistency function as a one-step TrigFlow-DDIM step from $t$ to $0$:
\begin{equation}
  \fth(x_t, t) = \cos(t)\,x_t - \sin(t)\,\sd\,\Fth(x_t/\sd,\, t).
  \label{eq:cm_param}
\end{equation}
This automatically satisfies the boundary condition $\fth(x_0, 0) = x_0$ because $\sin(0)=0$.

\subsection{Continuous-time CM training}
\label{sec:background:ctcm}

The continuous-time limit of discrete consistency distillation \citep[Theorem~5 and Remark~6]{song2023consistency} gives the following training gradient, where $w(t)$ includes the constant and time-sampling factors:
\begin{equation}
  \nabla_\theta\,\mathbb{E}_{x_t, t}\!\left[w(t)\;\fth^\top(x_t,t)\;\frac{d\fthm(x_t,t)}{dt}\right],
  \label{eq:ctcm_gradient}
\end{equation}
where $\theta^-$ denotes a stop-gradient on $\theta$ (no separate EMA target network is used, following \citealt{song2024improved,lu2025simplifying}). Expanding the time derivative along the PF-ODE gives
\begin{equation}
  \frac{d\fthm}{dt} = \cos(t)\!\left(\frac{dx_t}{dt} - \sd\,\Fthm\right)
  - \sin(t)\!\left(x_t + \sd\,\frac{d\Fthm}{dt}\right),
  \label{eq:tangent}
\end{equation}
where the second term carries $d\Fthm/dt$, computed by a forward-mode Jacobian--vector product (JVP) and the source of training instability.

\subsection{sCM's six stabilization fixes}
\label{sec:background:fixes}

sCM identifies the instability in \eqref{eq:tangent} as:
\begin{equation}
  \sin(t)\,\partial_t\Fthm = \sin(t)\cdot\frac{\partial\cnoise}{\partial t}
  \cdot\frac{\partial\,\mathrm{emb}(\cnoise)}{\partial\cnoise}
  \cdot\frac{\partial\Fthm}{\partial\,\mathrm{emb}}.
  \label{eq:instability_decomp}
\end{equation}
Three architectural changes address these factors, and three further changes modify the training objective.

With logarithmic time conditioning, $\cnoise(t)\propto\log(\sd\tan t)$, the factor $\sin(t)\,\partial_t\cnoise\propto1/\cos(t)$ diverges as $t\to\pi/2$. sCM uses identity conditioning, $\cnoise(t)=t$, for which this factor is the bounded $\sin(t)$. The second change concerns the Fourier time embedding, whose large effective scale magnifies $\partial\,\mathrm{emb}/\partial\cnoise$. A positional embedding drops that scale to about $0.02$. The third change is Adaptive Double Normalization (ADN), which replaces AdaGN \citep{dhariwal2021diffusion} with
\begin{equation}
  y = \norm(y) \odot \pnorm(s(t)) + \pnorm(b(t)),
  \label{eq:adn}
\end{equation}
where $\norm$ is GroupNorm (DDPM++) or channel-wise RMS normalization (EDM2), and $\pnorm$ is pixel normalization (Section~\ref{sec:teacher:adn}). Capping the scale and shift vectors this way holds $\partial\Fthm/\partial\,\mathrm{emb}$ down throughout training.

The other three fixes act on the gradient itself. Tangent normalization divides the raw tangent by its own Euclidean norm plus a constant, $g\to g/(\|g\|_2+c)$ with $c=0.1$ (Section~\ref{sec:student:details}). Adaptive weighting rewrites the inner-product gradient \eqref{eq:ctcm_gradient} as an MSE, $\nabla_\theta\mathbb{E}[\Fth^\top y]=\tfrac{1}{2}\nabla_\theta\mathbb{E}[\|\Fth-\Fthm+y\|^2]$, and trains a log-variance head $w_\phi(t)$ next to $\Fth$. Finally, the tangent warmup ramps the coefficient $r=\min(1,\,\text{iter}/H)$ on the $\sin(t)$ term from $0$ to $1$ over the first $H=10{,}000$ iterations.

\subsection{Multistep Consistency Distillation}
\label{sec:background:mscd}

\citet{heek2024multistep} discretize the generation trajectory into $M$ segments with boundaries $0=t_0<t_1<\cdots<t_M$, train a shared network across segments, and use DDIM teacher steps within each segment. They propose the aDDIM correction to compensate for the variance shortfall of finite DDIM steps. The paper trains on the discrete grid. It also notes that continuous-time training is possible provided the model takes an extra conditioning signal that identifies the current segment (so the prediction can change discontinuously across segment boundaries), and reports that early experiments with this variant worked comparably (\citealp{heek2024multistep}, \S3). Our MS-sCD (Section~\ref{sec:msscd}) builds on this approach by adding a segment-boundary input $\tk$ to the network and training in continuous time within the TrigFlow CM framework.

\section{TrigFlow Teacher: Architecture and Implementation}
\label{sec:teacher}

\subsection{Backbones}
\label{sec:teacher:arch}

We train at two settings, matching sCM's own backbone choices. CIFAR-10 \citep{krizhevsky2009cifar} (unconditional, $32{\times}32$) uses the Score~SDE DDPM++ network of \citet{song2021score} with the EDM~2022 \citep{karras2022elucidating} training settings. ImageNet-64 (class-conditional, $64{\times}64$) is a downsampled version of ImageNet \citep{deng2009imagenet,russakovsky2015imagenet} and uses the magnitude-preserving EDM2 UNet \citep{karras2024analyzing}. Both carry the TrigFlow parameterization, identity time conditioning, positional time embeddings, and ADN (Section~\ref{sec:background:fixes}). The EDM2 backbone is built from magnitude-preserving convolutions and uses EDM2's \texttt{normalize} ($\varepsilon=10^{-4}$) for the pixel normalization inside ADN. DDPM++ has no weight-normalized layers and instead applies the PGGAN pixel norm \citep{karras2017progressive} ($\varepsilon=10^{-8}$) that sCM cites. The rest of the section covers implementation choices that the paper leaves unspecified.

\subsection{ADN block and parameter count}
\label{sec:teacher:adn_variant}

ADN modulates every residual block as $y \cdot \pnorm(s) + \pnorm(b)$, with the scale $s$ and shift $b$ produced by a single linear projection of the time embedding. Relative to EDM2's scale-only modulation $y \cdot (c + 1)$, this scale-and-shift form requires the time-conditioning projection in each block to emit $2C$ channels instead of $C$, doubling the projection's weight tensor and adding roughly $C \cdot d_{\text{emb}}$ parameters per block, where $d_{\text{emb}}$ is the time-embedding dimension. Summed over the encoder and decoder blocks of the UNet, this accounts for the parameter-count gap we observe between the size reported by sCM ($280.2$M for size~S, the EDM2-S baseline figure) and our re-implementation, which has $293{,}922{,}753$ parameters (${\approx}294$M, about $5\%$ larger). The same offset is expected at the larger model sizes.

\subsection{Adaptive-weighting head and loss form}
\label{sec:teacher:weight_head}

The paper does not specify the architecture of $w_\phi(t)$. We use a fixed positional embedding (128 frequencies on $t$) followed by a single magnitude-preserving linear map to a scalar, giving one $w_\phi(t)$ per image.

For the loss, we minimize the EDM2 form
\begin{equation}
  \ell_{\text{ours}}(x_0, z, t) = \frac{\|\sd\Fth(x_t/\sd, t) - v_t\|^2}{e^{w_\phi(t)}} + D\,w_\phi(t),
  \label{eq:edm2_weight}
\end{equation}
with $D=C\!\times\!H\!\times\!W$ entries per image, summed over entries and averaged over the batch. The sCM appendix writes the corresponding scalar-per-image objective as
\begin{equation}
  \ell_{\text{sCM}}(x_0, z, t) = \frac{e^{w_\phi(t)}}{D}\,\|\sd\Fth(x_t/\sd, t) - v_t\|^2 - w_\phi(t).
  \label{eq:scm_weight}
\end{equation}

\subsection{Pixel-normalization}
\label{sec:teacher:adn}

For the EDM2 backbone, we implement $\pnorm$ with EDM2's \texttt{normalize} (pixel normalization in the sense of \citealt{karras2017progressive}), applied along the channel dimension:
\begin{equation}
  \pnorm(v) = \frac{v}{\mathrm{RMS}(v) + \varepsilon}, \quad \varepsilon = 10^{-4},
  \label{eq:normalize}
\end{equation}
for a length-$C$ channel vector $v$. Appendix~\ref{app:adn} restates the full routine. The teacher training details and other hyperparameter choices are in Appendix~\ref{app:experimental_details}.

\section{sCM Student: sCD and sCT}
\label{sec:student}

\subsection{Loss formulation}
\label{sec:student:loss}

The student is initialized from the teacher checkpoint. The loss minimized during sCD training is:
\begin{equation}
  \mathcal{L}_{\text{sCD}}(\theta,\phi) = \mathbb{E}_{x_0,z,t}\!\left[
  \frac{\|{\Fth - \Fthm - g}\|_2^2}{D\,e^{w_\phi(t)}} + D\,w_\phi(t)
  \right],
  \label{eq:scd_loss}
\end{equation}
where $D$ is the number of entries per image, $\Fthm$ denotes the student forward pass with stop-gradient, and $g$ is the normalized tangent (Section~\ref{sec:student:details}). The sCD teacher provides $dx_t/dt = \sd\,F_{\text{teacher}}(x_t/\sd, t)$. For sCT, we replace the teacher prediction with the sampled conditional velocity $\cos(t)\,z - \sin(t)\,x_0$, whose conditional expectation given $(x_t,t)$ is the marginal flow velocity. The corresponding continuous-time consistency training pseudo-objective is given in \citet[Remark~10]{song2023consistency}.

The raw tangent vector before normalization is:
\begin{equation}
  \tilde{g} = -\cos^2(t)\!\left(\sd\,\Fthm - \frac{dx_t}{dt}\right)
  - r\cdot\cos(t)\sin(t)\!\left(x_t + \sd\,\frac{d\Fthm}{dt}\right),
  \label{eq:raw_g}
\end{equation}
where $r = \min(1,\,\text{iter}/H)$ is the tangent warmup ramp and $d\Fthm/dt$ is computed via JVP (Section~\ref{sec:student:details}).

\subsection{Computing the tangent}
\label{sec:student:details}

\textbf{(a) JVP input tangents.} The naive tangent $(dx_t/dt,\,1)$ overflows in FP16 near $t\to 0$ and $t\to\pi/2$, so sCM rescales it by $\cos(t)\sin(t)$. We feed the network the unscaled $x_t$ and let it apply its own $\cin=1/\sd$, with tangents
\begin{equation}
  v_x = \cos(t)\sin(t)\,\frac{dx_t}{dt}, \qquad v_t = \cos(t)\sin(t).
  \label{eq:jvp_tangents}
\end{equation}

\textbf{(b) Tangent normalization.} We normalize the tangent exactly as in sCM's Algorithm~1, $g \to g/(\|g\|_2 + c)$ with $c=0.1$, where $\|g\|_2$ is the Euclidean norm over each sample's $C\times H\times W$ entries.

\textbf{(c) In-place weight renormalization and the JVP.} We disable in-place weight renormalization during JVP calls (Appendix~\ref{app:experimental_details}). For the attention layers, we follow the EDM2 release, which notes that an optimized fused scaled-dot-product implementation had not been verified to produce identical results, and use a plain einsum implementation. The JVP rule for the optimized path in \citet{lu2025simplifying}'s appendix is therefore not used in our setup.

\subsection{sCD vs.\ sCT}
\label{sec:student:scd_vs_sct}

The two variants share the loss in \eqref{eq:scd_loss} and differ only in where the tangent $dx_t/dt$ comes from, which we gave in Section~\ref{sec:student:loss}. sCD requires two forward passes per training step, while sCT requires one. We initialize the student from the teacher's weights in both cases. This is easy to overlook for sCT, which is usually described as teacher-free. Algorithm~1 of \citet{lu2025simplifying} sets $\theta\leftarrow\theta_{\text{pretrain}}$, and the appendix says so directly, so the weights themselves still come from the diffusion teacher. That is the real difference from the train-from-scratch consistency training of \citet{song2023consistency}.

\section{Reproduction Results}
\label{sec:results}

\begin{table}[t]
  \centering
  \caption{ \textbf{TrigFlow teacher FID, 50k samples, standard ADM evaluation.} ImageNet-64 is class-conditional with the EDM2-S backbone. CIFAR-10 is unconditional with the DDPM++ backbone. The TrigFlow sampler is DPM-Solver-2S \citep{lu2022dpm}, with Heun's intermediate step. }
  \label{tab:teacher}
  \begin{tabular}{llcc}
    \toprule
    Dataset & Method & NFE & FID \\
    \midrule
    ImageNet-64 & EDM2-S \citep{karras2024analyzing} & 63 & 1.58 \\
    ImageNet-64 & TrigFlow teacher \citep{lu2025simplifying} & 63 & 1.70 \\
    ImageNet-64 & TrigFlow teacher (ours) & 63 & \textbf{1.83} \\
    \midrule
    CIFAR-10 & TrigFlow teacher \citep{lu2025simplifying} & 35 & 2.15 \\
    CIFAR-10 & TrigFlow teacher (ours) & 35 & \textbf{2.08} \\
    \bottomrule
  \end{tabular}
\end{table}

\begin{table}[t]
  \centering
  \caption{ \textbf{sCD and sCT results on ImageNet-64 (EDM2-S), without classifier-free guidance (CFG).} }
  \label{tab:students}
  \begin{tabular}{lcccc}
    \toprule
    Method & Backbone & NFE=1 FID & NFE=2 FID & Framework \\
    \midrule
    iCT-deep \citep{song2024improved} & ADM-style & 3.25 & 2.77 & discrete, EDM \\
    ECM-S\textsuperscript{*} (ECT) \citep{geng2024consistency} & EDM2-S & 4.05 & 2.79 & discrete, EDM \\
    MSCD$^{\dagger}$ \citep{heek2024multistep} & UViT-small & 4.3 & 2.0 & discrete \\
    \midrule
    sCM sCD-S$^{\ddagger}$ \citep{lu2025simplifying} & EDM2-S/TrigFlow & 2.97 & 2.07 & TrigFlow \\
    sCM sCT-S$^{\ddagger}$ \citep{lu2025simplifying} & EDM2-S/TrigFlow & 3.23 & 2.93 & TrigFlow \\
    \midrule
    Ours sCD & EDM2-S/TrigFlow & 3.59 & 2.66 & TrigFlow \\
    Ours sCT & EDM2-S/TrigFlow & 4.34 & 3.98 & TrigFlow \\
    \bottomrule
  \end{tabular}
  \vspace{2pt}

  {\small \textsuperscript{*}102.4M tuning images. $^{\dagger}$ Heek et al.'s MSCD uses a UViT backbone (transformer UNet with 16$\times$1024-channel mid-blocks at the middle stage), structurally different from the magnitude-preserving EDM2 UNet we use. That model is also substantially larger in parameter count than the EDM2-S/TrigFlow models compared here, so its FID values are given for context rather than as an architecture-matched baseline. $^{\ddagger}$ Numbers from sCM Table~8 (appendix), at the EDM2-S size matched to our setup. The main-paper sCM table reports the XL variant for ImageNet-64.}
\end{table}

\begin{table}[t]
  \centering
  \caption{ \textbf{sCD and sCT results on unconditional CIFAR-10, DDPM++ backbone, 50k samples.} sCM paper numbers from \citet{lu2025simplifying}, Table~1. }
  \label{tab:students_cifar}
  \begin{tabular}{lccc}
    \toprule
    Method & NFE=1 FID & NFE=2 FID & Framework \\
    \midrule
    iCT \citep{song2024improved} & 2.83 & 2.46 & discrete, EDM \\
    ECT \citep{geng2024consistency} & 3.60 & 2.11 & discrete, EDM \\
    CD-LPIPS \citep{song2023consistency} & 3.55 & 2.93 & discrete, EDM \\
    \midrule
    sCM sCD \citep{lu2025simplifying} & 3.66 & 2.52 & TrigFlow \\
    sCM sCT \citep{lu2025simplifying} & 2.85 & 2.06 & TrigFlow \\
    \midrule
    Ours sCD & \textbf{3.59} & \textbf{2.39} & TrigFlow \\
    Ours sCT & 2.88 & 2.09 & TrigFlow \\
    \bottomrule
  \end{tabular}
\end{table}

Table~\ref{tab:teacher} reports the TrigFlow teacher FIDs, and the two datasets part ways already at this stage. On CIFAR-10, we reproduce the reported teacher performance. On ImageNet-64 it reaches $1.83$ against the reported $1.70$. We followed the appendix schedule in full and swept the post-hoc phEMA $\sigma_{\text{rel}}$, settling on $0.115$ rather than the prescribed $0.075$, which narrowed the ImageNet gap without closing it.

Tables~\ref{tab:students} and~\ref{tab:students_cifar} report our sCD and sCT results on ImageNet-64 and CIFAR-10. On CIFAR-10, where the method is fully specified, both students match the reported results, giving us confidence that our training and evaluation are implemented correctly. ImageNet-64 is the harder case. Despite training for the full prescribed schedule, our sCD-S trails the reported size-S numbers by about $0.6$ FID at both step counts ($3.59$ vs.\ $2.97$ at NFE${=}1$, and $2.66$ vs.\ $2.07$ at NFE${=}2$), and sCT-S trails by more. We discuss this further in Section~\ref{sec:discussion:backbone_gap}.

\section{Extension: Continuous-Time Multistep sCD (MS-sCD)}
\label{sec:msscd}

\subsection{Motivation}
\label{sec:msscd:motivation}

sCM's two-step sampling uses a fixed intermediate time $t_{\text{mid}} = 1.1$, whose optimal value varies with model size and training duration. More broadly, sCM's multi-step generation applies a single-step CM repeatedly, whereas MSCD trains a shared network across segments and runs one step per segment (Figure~\ref{fig:msscd_concept}). The sCM authors list ``incorporating our proposed techniques into multi-step consistency models'' as future work, and \citet{heek2024multistep} note that continuous-time training of an MSCD-style model is possible if the network takes an explicit conditioning signal identifying the current segment, so that the prediction can change discontinuously across segment boundaries.

MS-sCD combines these two ideas. We add a segment-boundary input $\tk$ to the sCM network through a zero-initialized embedding, so the model reduces to sCM at $M=1$ and becomes a segment-conditioned multistep CM otherwise. Three things change as a result: the segment CM parameterization (Section~\ref{sec:msscd:param}), its JVP target (Section~\ref{sec:msscd:loss}), and the training procedure (Section~\ref{sec:msscd:algorithm}).

\subsection{Segment-conditioned TrigFlow CM}
\label{sec:msscd:param}

We discretize $[0, \tmax]$ into $M$ segments with boundaries $0 = t_0 < t_1 < \cdots < t_M = \tmax$, $\tmax = \arctan(\sigma_{\max}/\sd)$. We fix the segment boundaries and space them uniformly in TrigFlow time $t$. Segment $k$ covers $(t_k, t_{k+1}]$. For a point $x_t$ in segment $k$, define $\dt := t - \tk$. The \emph{segment CM} is:

\begin{equation}
  f^{(k)}_\theta(x_t, t) := \cos(\dt)\,x_t - \sin(\dt)\,\sd\,\Fth\!\left(\frac{x_t}{\sd}, t, \tk\right).
  \label{eq:msscd_cm}
\end{equation}

\begin{figure}[t]
  \centering
  \begin{subfigure}[t]{0.48\textwidth}
    \centering
    \includegraphics[width=\linewidth]{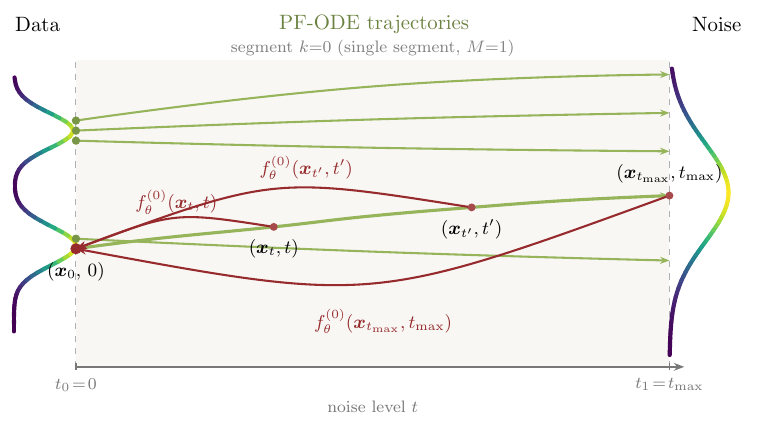}
    \caption{$M=1$ (sCM): a single segment covers $[0, t_{\max}]$. $f^{(0)}_\theta$ maps any point on a PF-ODE trajectory to the data origin $(x_0,0)$.}
    \label{fig:msscd_concept_m1}
  \end{subfigure}\hfill
  \begin{subfigure}[t]{0.48\textwidth}
    \centering
    \includegraphics[width=\linewidth]{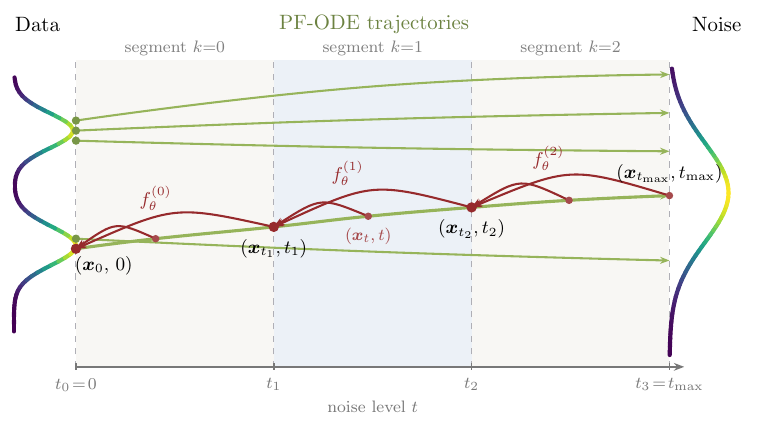}
    \caption{$M=3$ (MS-sCD): three segments. $f^{(k)}_\theta$ maps any point $(x_t, t)$ inside segment $k$ to that segment's lower boundary $(x_{\tk}, \tk)$.}
    \label{fig:msscd_concept_m3}
  \end{subfigure}
  \caption{ Green arrows are PF-ODE trajectories (data $\leftrightarrow$ noise) and red arrows show the consistency map. \textbf{(a)} For $M=1$ the segment CM \eqref{eq:msscd_cm} reduces to sCM: $f^{(0)}_\theta$ collapses every trajectory point to the data origin $(x_0, 0)$. \textbf{(b)} For $M>1$, MS-sCD splits the time axis at $0 = t_0 < t_1 < \cdots < t_M = t_{\max}$ and trains one segment-conditioned map $f^{(k)}_\theta$ per segment $k$. Each $f^{(k)}_\theta$ collapses trajectory points inside segment $k$ to the segment's lower boundary $(x_{\tk}, \tk)$.}
  \label{fig:msscd_concept}
\end{figure}

This is a TrigFlow-DDIM step from $t$ to $\tk$. It satisfies the \emph{segment boundary condition}: $f^{(k)}_\theta(x_{\tk}, \tk) = x_{\tk}$ because $\sin(0)=0$, so the CM is the identity at the bottom of each segment. Figure~\ref{fig:msscd_concept} illustrates the parameterization.

The network $\Fth$ takes $\tk$ as an additional conditioning input, implemented as:
\begin{equation}
  \mathrm{emb} \;\mathrel{+}=\; \text{linear}_{\text{init}=0}\!\left(\mathrm{PositionalEmbed}(\tk)\right),
  \label{eq:tk_embed}
\end{equation}
with a zero-initialized linear projection.
\subsection{MS-sCD loss}
\label{sec:msscd:loss}

We apply the continuous-time distillation pseudo-objective of \citet[Theorem~5 and Remark~6]{song2023consistency} to the segment CM, following sCM. Since $\tk$ is constant along any PF-ODE trajectory, it does not enter the JVP tangent.

The total time derivative of $f^{(k)}_{\theta^-}$ along the ODE is:
\begin{equation}
  \frac{df^{(k)}_{\theta^-}}{dt} =
  \cos(\dt)\!\left(\frac{dx_t}{dt} - \sd\,\Fthm\right)
  - \sin(\dt)\!\left(x_t + \sd\,\frac{d\Fthm}{dt}\right).
  \label{eq:msscd_tangent}
\end{equation}

After the inner-product-to-MSE conversion and prior weighting $\lambda(t,k) = 1/(\sd\tan\dt)$:
\begin{equation}
  \mathcal{L}_{\text{MS-sCD}}(\theta,\phi) = \mathbb{E}\!\left[
  \frac{1}{D\,e^{w_\phi(t)}}\left\|
  \Fth - \Fthm - g_k
  \right\|_2^2 + D\,w_\phi(t)
  \right],
  \label{eq:msscd_loss}
\end{equation}
where $g_k=\tilde g_k/(\|\tilde g_k\|_2+c)$, with raw target
\begin{equation}
  \tilde{g}_k = -\cos^2(\dt)\!\left(\sd\,\Fthm - \frac{dx_t}{dt}\right)
  - r\cdot\cos(\dt)\sin(\dt)\!\left(x_t + \sd\,\frac{d\Fthm}{dt}\right).
  \label{eq:msscd_raw_g}
\end{equation}
The JVP tangent vectors are:
\begin{equation}
  v_x = \cos(\dt)\sin(\dt)\cdot dx_t/dt, \qquad v_t = \cos(\dt)\sin(\dt),
  \label{eq:msscd_jvp_tangents}
\end{equation}
and $\tk$ is passed as a primal input with zero tangent (since $d\tk/dt = 0$ along the PF-ODE).

\subsection{Training algorithm}
\label{sec:msscd:algorithm}

Algorithm~\ref{alg:msscd_train} is Algorithm~1 of \citet{lu2025simplifying} with three changes. Namely, a segment lookup that sets $\tk$ from the segment $t$ falls into, the substitution $\dt = t-\tk$ in the tangent, and the extra conditioning input $\tk$ passed to the network on both the differentiated and the stop-gradient branches of the loss. At $M=1$, we set $t_0=0$, $t_1=\tmax$, so $\dt = t$ for all $t \in (0, \tmax]$. Then \eqref{eq:msscd_loss} reduces exactly to $\mathcal{L}_{\text{sCD}}$ \eqref{eq:scd_loss}.

\begin{algorithm}[t]
  \caption{Continuous-time Multistep sCD (MS-sCD) training, sCD variant.}
  \label{alg:msscd_train}
  \begin{algorithmic}[1]
    \STATE \textbf{Input:} data $\mathcal{D}$ (std.\ $\sd$), teacher $F_{\text{pretrain}}$,
    student $\Fth$ (init.\ from $F_{\text{pretrain}}$, $\tk$-embedding init.\ to zero),
    weighting $w_\phi$, segments $0=t_0<\cdots<t_M=\tmax$,
    proposal $(P_{\text{mean}},P_{\text{std}})$, norm const $c=0.1$, warmup $H$, lr $\eta$.
    \STATE \textbf{Init:} $\theta\leftarrow\theta_{\text{pretrain}}$, $\text{iter}\leftarrow 0$.
    \REPEAT
    \STATE $x_0\sim\mathcal{D}$,\; $z\sim\mathcal{N}(0,\sd^2 I)$
    \STATE $\tau\sim\mathcal{N}(P_{\text{mean}},P_{\text{std}}^2)$,\; $t\leftarrow\arctan(e^\tau/\sd)$
    \STATE find $k$ s.t.\ $t\in(t_k,t_{k+1}]$;\; $\dt\leftarrow t-t_k$
    \hfill\COMMENT{segment lookup}
    \STATE $x_t\leftarrow\cos(t)\,x_0+\sin(t)\,z$
    \STATE $dx_t/dt\leftarrow\sd\,F_{\text{pretrain}}(x_t/\sd,\,t)$
    \hfill\COMMENT{sCT: $\cos(t)z-\sin(t)x_0$}
    \STATE $r\leftarrow\min(1,\text{iter}/H)$ \hfill\COMMENT{tangent warmup}
    \STATE $Y\leftarrow\text{JVP}\big[\Fthm(\cdot,\cdot,t_k)\big]$ at $(x_t,\,t)$
    with tangent $\big(\cos\dt\sin\dt\,(dx_t/dt),\ \cos\dt\sin\dt\big)$
    \hfill\COMMENT{network applies $\cin{=}1/\sd$; $\dt=t-t_k$}
    \STATE $\tilde{g}\leftarrow-\cos^2\!\dt\,(\sd\Fthm-dx_t/dt)-r\,(\cos\dt\sin\dt\,x_t+\sd\,Y)$
    \STATE $g_k\leftarrow\tilde{g}/(\|\tilde{g}\|_2+c)$ \hfill\COMMENT{L2 tangent norm}
    \STATE $\mathcal{L}\leftarrow\frac{1}{D\,e^{w_\phi(t)}}\big\|\Fth(x_t/\sd,t,t_k)-\Fthm-g_k\big\|_2^2+D\,w_\phi(t)$
    \STATE $(\theta,\phi)\leftarrow(\theta,\phi)-\eta\nabla_{\theta,\phi}\mathcal{L}$;\;
    $\text{iter}\leftarrow\text{iter}+1$
    \UNTIL{convergence}
  \end{algorithmic}
\end{algorithm}

\subsection{Sampling}
\label{sec:msscd:sampling}

\begin{algorithm}[t]
  \caption{MS-sCD sampling ($M$ steps, $M$ NFE)}
  \label{alg:msscd_sample}
  \begin{algorithmic}[1]
    \STATE Draw $x_{t_M} \sim \mathcal{N}(0, \sd^2 I)$
    \FOR{$k = M-1, M-2, \ldots, 0$}
    \STATE $\Delta_k \leftarrow t_{k+1} - t_k$
    \STATE $x_{t_k} \leftarrow \cos(\Delta_k)\,x_{t_{k+1}}
    - \sin(\Delta_k)\,\sd\,\Fth\!\left(x_{t_{k+1}}/\sd,\; t_{k+1},\; t_k\right)$
    \ENDFOR
    \RETURN $x_{t_0}$
  \end{algorithmic}
\end{algorithm}

Algorithm~\ref{alg:msscd_sample} gives the $M$-step sampling procedure. Each step applies the segment CM \eqref{eq:msscd_cm} from $t_{k+1}$ to $t_k$, for a total cost of $M$ network evaluations. At $M=2$, the NFE budget matches sCM's two-step sampling, with the intermediate time $t_{\text{mid}}$ replaced by the segment boundary $t_1$ and the predict-and-re-noise procedure replaced by deterministic segment maps.

\subsection{Results}
\label{sec:msscd:results}

\begin{table}[t]
  \centering
  \caption{ \textbf{MS-sCD vs.\ baselines on ImageNet-64 (EDM2-S, class-conditional, no CFG, 50k samples).} sCM-S baselines are from \citet{lu2025simplifying}, appendix Table~8, where the main-paper sCM table reports the XL variant for ImageNet-64. The MSCD baselines on the EDM2-S backbone are \emph{our} reproductions of \citet{heek2024multistep}'s method on this architecture. }
  \label{tab:msscd}
  \begin{tabular}{lccc}
    \toprule
    Method & $M$ & NFE & FID \\
    \midrule
    sCM sCD-S \citep{lu2025simplifying} (appendix Table~8) & 1 & 1 & 2.97 \\
    sCM sCD-S \citep{lu2025simplifying} (appendix, 2-step) & 1 & 2 & 2.07 \\
    \midrule
    MSCD on EDM2-S (ours) & 2 & 2 & 3.44 \\
    MSCD on EDM2-S (ours) & 4 & 4 & 2.32 \\
    MSCD on EDM2-S (ours) & 8 & 8 & 1.79 \\
    \midrule
    MS-sCD (ours) & 2 & 2 & 2.53 \\
    MS-sCD (ours) & 4 & 4 & 2.26 \\
    MS-sCD (ours) & 8 & 8 & 2.08 \\
    \midrule
    TrigFlow teacher (ours) & --- & 63 & 1.83 \\
    \bottomrule
  \end{tabular}
\end{table}

\begin{figure}[t]
  \centering
  \includegraphics[width=\linewidth]{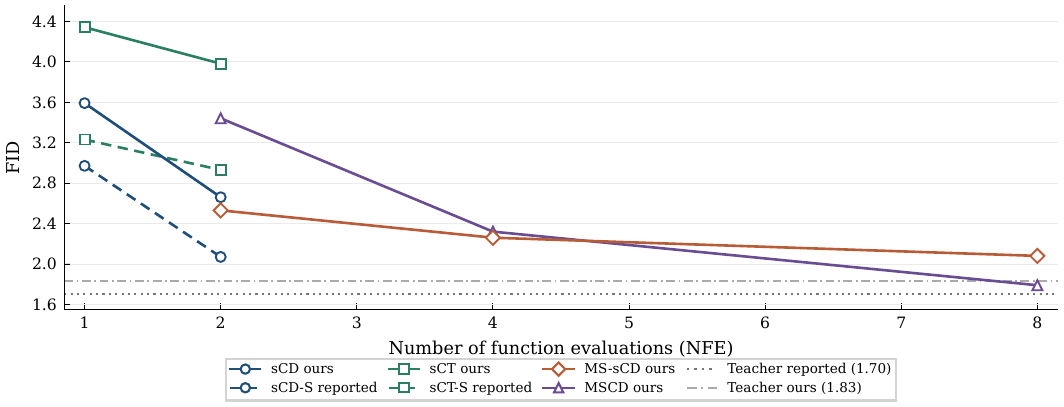}
  \caption{FID vs.\ NFE on ImageNet-64 (EDM2-S, class-conditional, no CFG, 50k samples). Solid lines: our EDM2-S/TrigFlow reproduction. Dashed lines: sCM's reported sCD-S and sCT-S numbers \citep[appendix Table~8]{lu2025simplifying}. The two multistep students extend past the $\text{NFE}{\le}2$ regime of the single-step CMs by varying the number of segments $M\in\{2,4,8\}$: continuous MS-sCD (orange) leads at $M{=}2$, while our discrete MSCD reproduction (purple) overtakes it by $M{=}8$, both approaching the teacher reference (horizontal lines: ours $1.83$, sCM-reported $1.70$).}
  \label{fig:fid_vs_m}
\end{figure}

Table~\ref{tab:msscd} reports MS-sCD FID at $M=2,4,8$ against MSCD and the sCM baseline. At $M{=}2$, MS-sCD outperforms our discrete-time MSCD reproduction by $0.91$ FID ($2.53$ vs.\ $3.44$), consistent with the continuous-time formulation avoiding the discretization error that finite DDIM steps incur in the low-NFE regime. MSCD closes the gap at larger $M$ and slightly surpasses MS-sCD at $M{=}8$ ($1.79$ vs.\ $2.08$), approaching the teacher FID. Figure~\ref{fig:fid_vs_m} shows FID vs.\ NFE, with MS-sCD steadily decreasing toward the teacher FID as $M$ grows.

Table~\ref{tab:msscd_cifar} reports the same $M\in\{2,4,8\}$ sweep on unconditional CIFAR-10. FID improves with $M$ ($2.60$, $2.35$, $2.20$), matching a 199-NFE teacher evaluation within seed variation (FID $2.22$), while remaining above our two-step sCT ($2.09$) and the 35-NFE teacher used in the main tables ($2.08$). We discuss this comparison in Appendix~\ref{app:msscd_cifar}.

\begin{table}[t]
  \centering
  \caption{ \textbf{MS-sCD on unconditional CIFAR-10 (DDPM++ backbone, 50k samples).} Teacher rows use DPM-Solver-2S on our TrigFlow teacher; the 35-NFE entry matches Table~\ref{tab:teacher}. Two-step sCT is from Table~\ref{tab:students_cifar}. }
  \label{tab:msscd_cifar}
  \begin{tabular}{lccc}
    \toprule
    Method & $M$ & NFE & FID \\
    \midrule
    MS-sCD (ours) & 2 & 2 & 2.60 \\
    MS-sCD (ours) & 4 & 4 & 2.35 \\
    MS-sCD (ours) & 8 & 8 & 2.20 \\
    \midrule
    Ours sCT & 1 & 2 & 2.09 \\
    TrigFlow teacher (ours) & --- & 35 & 2.08 \\
    TrigFlow teacher (ours) & --- & 199 & 2.22 \\
    \bottomrule
  \end{tabular}
\end{table}

This is not a strict continuous-versus-discrete ablation: each student spaces its boundaries uniformly in its teacher's native schedule (the EDM/Karras $\sigma$-grid for MSCD, with the $M{=}2$ intermediate segment boundary at $\sigma=1.1$; TrigFlow $t$ for MS-sCD), and the two also differ in noise policy and checkpoint selection (Appendices~\ref{app:noise_policy} and~\ref{app:experimental_details}). In Appendix~\ref{app:matched_grid} we retrain MSCD on MS-sCD's boundary locations via the EDM--TrigFlow map $\sigma = \sd\tan(t)$; MSCD gets worse under that matched schedule, as expected when it is pulled off the teacher's native grid. Appendix~\ref{app:iter_scd} compares dedicated segment-conditioned training to simply rolling out the one-step student. Here, \emph{Song-style iterative sampling} means alternating clean-data prediction with fresh Gaussian re-noising, following the multistep consistency sampler of \citet[Algorithm~1]{song2023consistency}. Evaluating our sCD checkpoint with this procedure at NFE${=}4$ and $8$ shows that MS-sCD remains better at the same budget.

\section{Supporting Context: MSCD and Moment Matching}
\label{sec:context}

Before training the TrigFlow models we reproduced two existing distillation methods on class-conditional ImageNet-64, using the EDM framework \citep{karras2022elucidating} with the ADM-style UNet of \citet{dhariwal2021diffusion}, to validate our 50k-sample FID pipeline against published numbers. Our Multistep Consistency Distillation (MSCD, \citealt{heek2024multistep}) student reaches FID ${\approx}2.3$ at 8 steps, matching the deterministic Heun teacher at 511 NFE. This is the same method we later port to the EDM2-S backbone as the discrete-time baseline for MS-sCD (Table~\ref{tab:msscd}). Because MSCD trains the student to follow the deterministic teacher ODE within each segment, it is expected that the teacher quality has an influence on its FID. We separately reproduced moment matching \citep{salimans2024multistep}, a distribution-matching distillation, on the same backbone, reaching FID ${\approx}1.4$ with an 8-step student. We do not pursue this family further. The contrast is worth noting all the same. On images, distribution-matching distillation reaches much lower few-step FID than the trajectory-matching consistency objectives we focus on.

\section{Discussion}
\label{sec:discussion}

\subsection{The teacher gap propagates to the distilled students}
\label{sec:discussion:teacher_gap}

Our TrigFlow teacher reaches FID $1.83$ on ImageNet-64 against the reported $1.70$ (Table~\ref{tab:teacher}). Since sCD distills the teacher ODE, teacher quality may contribute to our sCD-S shortfall. The sCT-S gap, which does not distill the teacher, must have a separate cause (Section~\ref{sec:discussion:sct_scd}). The teacher deficit also reframes the multistep route: a one-step student must absorb the entire teacher-to-data map in a single evaluation, whereas MSCD and MS-sCD divide it across $M$ segments and lower the per-step burden, tracking the teacher more closely at the same total compute. We therefore expect that, with the residual teacher gap closed, the multistep budget--quality trade-off is the more dependable use of a fixed sampling budget at small model scale.

\subsection{sCT vs.\ sCD ordering depends on model capacity}
\label{sec:discussion:sct_scd}

\citet{lu2025simplifying} state in their Discussions and Limitations section that ``sCT consistently outperforms sCD on CIFAR-10 and ImageNet $64{\times}64$ but is less effective than sCD across different model scales on ImageNet $512{\times}512$,'' attributing the ImageNet-$512$ exception to the training variance of CT in latent spaces with imperfect encoders. The same paper's appendix table at ImageNet-$64$ (sCM Table~8) reports sCT-S at FID $3.23$/$2.93$ (1-/2-step) versus sCD-S at $2.97$/$2.07$, so sCT is worse than sCD at the smallest model size by ${\approx}0.26$ at NFE${=}1$ and ${\approx}0.86$ at NFE${=}2$. From size~M upward the ordering reverses (sCT-M $2.25$/$1.81$ vs.\ sCD-M $2.79$/$1.89$), and sCT wins through XL as the paper's claim suggests. The blanket statement therefore holds only at the larger scales. At EDM2-S the ordering is reversed even in a pixel-space setting where the latent-space hypothesis does not apply. The sCT/sCD ordering is therefore capacity-dependent, not purely dataset-dependent.

\subsection{Backbone-dependent reproduction quality on CIFAR-10 versus ImageNet-64}
\label{sec:discussion:backbone_gap}

On CIFAR-10 our sCD matches and slightly exceeds the reported FID at both step counts. On ImageNet-64, sCD-S lags by ${\approx}0.6$ FID and sCT-S by more. These results do not establish the cause of the gap. Two implementation choices specific to ImageNet-64 may contribute, but we have not tested them in controlled ablations.

\textbf{Adaptive-weighting head $w_\phi(t)$.} sCM specifies the weighting objective but not the architecture of $w_\phi$, only that it follows EDM2 (a Fourier embedding into a magnitude-preserving linear map). We use a positional embedding of $t$ instead, matching sCM's documented Fourier-to-positional substitution for the main network, which the paper does not extend explicitly to the weighting head.

\textbf{Attention implementation.} sCM states that the ImageNet-64 students use FlashAttention with a custom JVP rule, but the released EDM2 backbone uses an einsum attention by default (its notes flag the fused kernel as insufficiently tested). We follow that default. The FlashAttention path and its JVP rule cannot be reconstructed from the available artifacts, and the FP16 reduction-order difference could accumulate over the $400$k-iteration student schedule. Two architectural details are unspecified, and other undocumented implementation choices may also contribute to the gap.

\section{Conclusion and Future Work}
\label{sec:limitations}

We set out to reproduce the TrigFlow/sCM formulation and to see how far it extends. The three architectural changes address the terms in \eqref{eq:instability_decomp}, while the three objective changes modify the training gradient. On CIFAR-10, we recover the reported sCD FID at both step counts. On ImageNet-64, our teacher does not reach the reported $1.70$, and the students also fall short. Implementation differences may contribute to these gaps, but we have not isolated their effects.

Our MS-sCD results show that segment-conditioned distillation can be trained within the continuous-time sCD framework. FID improves as the number of segments increases on both datasets. On ImageNet-64, MS-sCD outperforms iterative sCD sampling at the same NFE, while its advantage over discrete-time MSCD is limited to the smaller step budgets. These findings are limited to pixel-space generation at $32{\times}32$ and $64{\times}64$ resolution. Evaluating the method at higher resolutions, in latent spaces, and for text-to-image generation is left to future work.

\subsubsection*{Acknowledgments} We thank the members of the Kumara and Lai labs for
their comments and feedback. We also thank Xiowen You, Ankur Verma, and Daniel Cullina for helpful discussions and suggestions, and Charles Pavloski (NCSA NSF-OAC2320345 \& NSF-OAC2005572) for feedback and troubleshooting. We gratefully acknowledge funding support from the Cornell University BRC Epigenomics Core Facility (RRID:SCR\_021287), the Penn State Institute for Computational and Data Sciences (RRID:SCR\_025154), and Penn State AIMI (RRID:SCR\_022867). Soundar R.\ Kumara would also like to thank the Allen E.\ Pearce and Allen M.\ Pearce Professorship. This project was supported by NIH grant R35GM155380 to W.K.M.L.\ and a gift to AIMI research from Dell Technologies. Computational resources were provided through an NSF ACCESS Maximize allocation (CIS260116).

\bibliography{tmlr}

@inproceedings{lu2025simplifying,
  title = {Simplifying, stabilizing and scaling continuous-time consistency models},
  author = {Lu, Cheng and Song, Yang},
  booktitle = {International Conference on Learning Representations (ICLR)},
  year = {2025},
  note = {Section, table, and algorithm references use arXiv:2410.11081v2},
  sourceurl = {https://arxiv.org/abs/2410.11081v2}
}

@article{heek2024multistep,
  title = {Multistep consistency models},
  author = {Heek, Jonathan and Hoogeboom, Emiel and Salimans, Tim},
  journal = {arXiv preprint arXiv:2403.06807},
  year = {2024},
  sourceurl = {https://arxiv.org/abs/2403.06807},
  doi = {10.48550/arXiv.2403.06807}
}

@inproceedings{salimans2024multistep,
  title = {Multistep distillation of diffusion models via moment matching},
  author = {Salimans, Tim and Mensink, Thomas and Heek, Jonathan and Hoogeboom, Emiel},
  volume = {37},
  pages = {36046--36070},
  year = {2024},
  booktitle = {Advances in Neural Information Processing Systems},
  sourceurl = {https://proceedings.neurips.cc/paper_files/paper/2024/hash/3f66d5cdbe032bb750f2dc523357b7a5-Abstract-Conference.html},
  doi = {10.52202/079017-1136}
}

@inproceedings{song2023consistency,
  title = {Consistency Models},
  author = {Song, Yang and Dhariwal, Prafulla and Chen, Mark and Sutskever, Ilya},
  booktitle = {Proceedings of the 40th International Conference on Machine Learning},
  volume = {202},
  series = {Proceedings of Machine Learning Research},
  pages = {32211--32252},
  publisher = {PMLR},
  year = {2023},
  sourceurl = {https://proceedings.mlr.press/v202/song23a.html}
}

@inproceedings{song2024improved,
  title = {Improved Techniques for Training Consistency Models},
  author = {Song, Yang and Dhariwal, Prafulla},
  booktitle = {International Conference on Learning Representations (ICLR)},
  year = {2024},
  sourceurl = {https://arxiv.org/abs/2310.14189}
}

@inproceedings{geng2024consistency,
  title = {Consistency models made easy},
  author = {Geng, Zhengyang and Pokle, Ashwini and Luo, Weijian and Lin, Justin and Kolter, J. Zico},
  booktitle = {International Conference on Learning Representations (ICLR)},
  year = {2025},
  sourceurl = {https://arxiv.org/abs/2406.14548}
}

@inproceedings{karras2022elucidating,
  title = {Elucidating the design space of diffusion-based generative models},
  author = {Karras, Tero and Aittala, Miika and Aila, Timo and Laine, Samuli},
  volume = {35},
  pages = {26565--26577},
  year = {2022},
  booktitle = {Advances in Neural Information Processing Systems},
  sourceurl = {https://proceedings.neurips.cc/paper_files/paper/2022/hash/a98846e9d9cc01cfb87eb694d946ce6b-Abstract.html},
  doi = {10.52202/068431-1926}
}

@inproceedings{karras2024analyzing,
  title = {Analyzing and Improving the Training Dynamics of Diffusion Models},
  author = {Karras, Tero and Aittala, Miika and Lehtinen, Jaakko and Hellsten, Janne and Aila, Timo and Laine, Samuli},
  year = {2024},
  booktitle = {Proceedings of the IEEE/CVF Conference on Computer Vision and Pattern Recognition},
  pages = {24174--24184},
  sourceurl = {https://openaccess.thecvf.com/content/CVPR2024/html/Karras_Analyzing_and_Improving_the_Training_Dynamics_of_Diffusion_Models_CVPR_2024_paper.html},
  doi = {10.1109/CVPR52733.2024.02282}
}

@inproceedings{dhariwal2021diffusion,
  title = {Diffusion Models Beat {GANs} on Image Synthesis},
  author = {Dhariwal, Prafulla and Nichol, Alexander},
  volume = {34},
  pages = {8780--8794},
  year = {2021},
  booktitle = {Advances in Neural Information Processing Systems},
  sourceurl = {https://proceedings.neurips.cc/paper/2021/hash/49ad23d1ec9fa4bd8d77d02681df5cfa-Abstract.html}
}

@inproceedings{salimans2022progressive,
  title = {Progressive Distillation for Fast Sampling of Diffusion Models},
  author = {Salimans, Tim and Ho, Jonathan},
  year = {2022},
  booktitle = {International Conference on Learning Representations (ICLR)},
  sourceurl = {https://arxiv.org/abs/2202.00512}
}

@inproceedings{song2021ddim,
  title = {Denoising Diffusion Implicit Models},
  author = {Song, Jiaming and Meng, Chenlin and Ermon, Stefano},
  booktitle = {International Conference on Learning Representations (ICLR)},
  year = {2021},
  sourceurl = {https://openreview.net/forum?id=St1giarCHLP}
}

@inproceedings{wang2023prolificdreamer,
  title = {{ProlificDreamer}: High-Fidelity and Diverse Text-to-{3D} Generation with Variational Score Distillation},
  author = {Wang, Zhengyi and Lu, Cheng and Wang, Yikai and Bao, Fan and Li, Chongxuan and Su, Hang and Zhu, Jun},
  booktitle = {Advances in Neural Information Processing Systems},
  year = {2023},
  volume = {36},
  pages = {8406--8441},
  sourceurl = {https://proceedings.neurips.cc/paper_files/paper/2023/hash/1a87980b9853e84dfb295855b425c262-Abstract-Conference.html},
  doi = {10.52202/075280-0368}
}

@inproceedings{yin2024dmd,
  title = {One-step Diffusion with Distribution Matching Distillation},
  author = {Yin, Tianwei and Gharbi, Micha\"el and Zhang, Richard and Shechtman, Eli and Durand, Fr\'edo and Freeman, William T. and Park, Taesung},
  booktitle = {Proceedings of the IEEE/CVF Conference on Computer Vision and Pattern Recognition},
  year = {2024},
  pages = {6613--6623},
  sourceurl = {https://openaccess.thecvf.com/content/CVPR2024/html/Yin_One-step_Diffusion_with_Distribution_Matching_Distillation_CVPR_2024_paper.html},
  doi = {10.1109/CVPR52733.2024.00632}
}

@inproceedings{lu2022dpm,
  title = {{DPM-Solver}: A Fast {ODE} Solver for Diffusion Probabilistic Model Sampling in Around 10 Steps},
  author = {Lu, Cheng and Zhou, Yuhao and Bao, Fan and Chen, Jianfei and Li, Chongxuan and Zhu, Jun},
  volume = {35},
  pages = {5775--5787},
  booktitle = {Advances in Neural Information Processing Systems},
  year = {2022},
  sourceurl = {https://proceedings.neurips.cc/paper_files/paper/2022/hash/260a14acce2a89dad36adc8eefe7c59e-Abstract-Conference.html},
  doi = {10.52202/068431-0418}
}

@inproceedings{ho2020denoising,
  title = {Denoising diffusion probabilistic models},
  author = {Ho, Jonathan and Jain, Ajay and Abbeel, Pieter},
  volume = {33},
  pages = {6840--6851},
  year = {2020},
  booktitle = {Advances in Neural Information Processing Systems},
  sourceurl = {https://proceedings.neurips.cc/paper/2020/hash/4c5bcfec8584af0d967f1ab10179ca4b-Abstract.html}
}

@inproceedings{sohl2015deep,
  title = {Deep unsupervised learning using nonequilibrium thermodynamics},
  author = {Sohl-Dickstein, Jascha and Weiss, Eric and Maheswaranathan, Niru and Ganguli, Surya},
  booktitle = {Proceedings of the 32nd International Conference on Machine Learning},
  pages = {2256--2265},
  year = {2015},
  volume = {37},
  series = {Proceedings of Machine Learning Research},
  publisher = {PMLR},
  sourceurl = {https://proceedings.mlr.press/v37/sohl-dickstein15.html}
}

@inproceedings{song2021score,
  title = {Score-Based Generative Modeling through Stochastic Differential Equations},
  author = {Song, Yang and Sohl-Dickstein, Jascha and Kingma, Diederik P and Kumar, Abhishek and Ermon, Stefano and Poole, Ben},
  booktitle = {International Conference on Learning Representations (ICLR)},
  year = {2021},
  sourceurl = {https://openreview.net/forum?id=PxTIG12RRHS}
}

@inproceedings{liu2022flow,
  title = {Flow straight and fast: Learning to generate and transfer data with rectified flow},
  author = {Liu, Xingchao and Gong, Chengyue and Liu, Qiang},
  year = {2023},
  booktitle = {International Conference on Learning Representations (ICLR)},
  sourceurl = {https://openreview.net/forum?id=XVjTT1nw5z}
}

@article{albergo2023stochastic,
  title = {Stochastic interpolants: A unifying framework for flows and diffusions},
  author = {Albergo, Michael S. and Boffi, Nicholas M. and Vanden-Eijnden, Eric},
  journal = {Journal of Machine Learning Research},
  volume = {26},
  number = {209},
  pages = {1--80},
  year = {2025},
  sourceurl = {https://jmlr.org/papers/v26/23-1605.html}
}

@inproceedings{karras2017progressive,
  title = {Progressive Growing of {GAN}s for Improved Quality, Stability, and Variation},
  author = {Karras, Tero and Aila, Timo and Laine, Samuli and Lehtinen, Jaakko},
  booktitle = {International Conference on Learning Representations (ICLR)},
  year = {2018},
  sourceurl = {https://openreview.net/forum?id=Hk99zCeAb}
}

@inproceedings{lipman2022flow,
  title = {Flow Matching for Generative Modeling},
  author = {Lipman, Yaron and Chen, Ricky T. Q. and Ben-Hamu, Heli and Nickel, Maximilian and Le, Matt},
  year = {2023},
  booktitle = {International Conference on Learning Representations (ICLR)},
  sourceurl = {https://openreview.net/forum?id=PqvMRDCJT9t}
}

@inproceedings{kynkaanniemi2019improved,
  title = {Improved precision and recall metric for assessing generative models},
  author = {Kynk{\"a}{\"a}nniemi, Tuomas and Karras, Tero and Laine, Samuli and Lehtinen, Jaakko and Aila, Timo},
  booktitle = {Advances in Neural Information Processing Systems},
  volume = {32},
  year = {2019},
  sourceurl = {https://proceedings.neurips.cc/paper/2019/hash/0234c510bc6d908b28c70ff313743079-Abstract.html}
}

@inproceedings{naeem2020reliable,
  title = {Reliable fidelity and diversity metrics for generative models},
  author = {Naeem, Muhammad Ferjad and Oh, Seong Joon and Uh, Youngjung and Choi, Yunjey and Yoo, Jaejun},
  booktitle = {Proceedings of the 37th International Conference on Machine Learning},
  pages = {7176--7185},
  year = {2020},
  volume = {119},
  series = {Proceedings of Machine Learning Research},
  publisher = {PMLR},
  sourceurl = {https://proceedings.mlr.press/v119/naeem20a.html}
}

@inproceedings{heusel2017fid,
  title = {{GANs} Trained by a Two Time-Scale Update Rule Converge to a Local {Nash} Equilibrium},
  author = {Heusel, Martin and Ramsauer, Hubert and Unterthiner, Thomas and Nessler, Bernhard and Hochreiter, Sepp},
  booktitle = {Advances in Neural Information Processing Systems},
  volume = {30},
  year = {2017},
  sourceurl = {https://proceedings.neurips.cc/paper/2017/hash/8a1d694707eb0fefe65871369074926d-Abstract.html}
}

@techreport{krizhevsky2009cifar,
  title = {Learning Multiple Layers of Features from Tiny Images},
  author = {Krizhevsky, Alex},
  institution = {University of Toronto},
  year = {2009},
  sourceurl = {https://www.cs.toronto.edu/~kriz/learning-features-2009-TR.pdf}
}

@inproceedings{deng2009imagenet,
  title = {{ImageNet}: A Large-Scale Hierarchical Image Database},
  author = {Deng, Jia and Dong, Wei and Socher, Richard and Li, Li-Jia and Li, Kai and Fei-Fei, Li},
  booktitle = {Proceedings of the IEEE Conference on Computer Vision and Pattern Recognition},
  pages = {248--255},
  year = {2009},
  sourceurl = {https://image-net.org/static_files/papers/imagenet_cvpr09.pdf},
  doi = {10.1109/CVPR.2009.5206848}
}

@article{russakovsky2015imagenet,
  title = {{ImageNet} Large Scale Visual Recognition Challenge},
  author = {Russakovsky, Olga and Deng, Jia and Su, Hao and Krause, Jonathan and Satheesh, Sanjeev and Ma, Sean and Huang, Zhiheng and Karpathy, Andrej and Khosla, Aditya and Bernstein, Michael and Berg, Alexander C. and Fei-Fei, Li},
  journal = {International Journal of Computer Vision},
  volume = {115},
  number = {3},
  pages = {211--252},
  year = {2015},
  doi = {10.1007/s11263-015-0816-y},
  sourceurl = {https://doi.org/10.1007/s11263-015-0816-y}
}

@misc{mathew2026software,
  author = {Mathew, Vinay Saji and Kumara, Soundar R. and Kellogg, Gretta D. and Lai, William KM},
  title = {Stable Continuous-Time Consistency Distillation: An Empirical Study with a Multistep Extension},
  year = {2026},
  howpublished = {\href{https://archive.softwareheritage.org/swh:1:dir:a63f1ad1c0e9e6746a3b31cf9dab13ecfa0b6ead;origin=https://github.com/EpiGenomicsCode/edm2_sCM;visit=swh:1:snp:efd62352415c203fdbb0ffa50b2ac63aa994b8c7;anchor=swh:1:rev:cb05b7b5a344868acd08b291fac956f15b710adf}{Software Heritage archive}},
  note = {Version 1.0. SWHID: \nolinkurl{swh:1:dir:a63f1ad1c0e9e6746a3b31cf9dab13ecfa0b6ead}}
}
\bibliographystyle{tmlr}

\appendix

\section{Derivations}
\label{app:trigflow_derivations}

\subsection{CM parameterization as a TrigFlow-DDIM step}

The deterministic DDIM update holds the inferred clean sample and noise fixed during each step \citep{song2021ddim}. In TrigFlow coordinates, let $\hat v_t=\sd F_\theta(x_t/\sd,t)$. Inverting the trigonometric relations between data, noise, and velocity gives
\begin{align}
  \hat x_0 &= \cos(t)\,x_t-\sin(t)\,\hat v_t, \\
  \hat z &= \sin(t)\,x_t+\cos(t)\,\hat v_t,
\end{align}
where $\hat z$ uses the same $\sd$-scaled noise convention as \eqref{eq:trigflow_process}. For $0\le s<t$, substituting these fixed estimates into the DDIM update yields
\begin{align}
  x_s^{\mathrm{DDIM}}
  &= \cos(s)\,\hat x_0+\sin(s)\,\hat z \nonumber\\
  &= \cos(t-s)\,x_t-\sin(t-s)\,\sd F_\theta(x_t/\sd,t).
  \label{eq:trigflow_ddim}
\end{align}
This is the first-order TrigFlow sampler described by \citet[Appendix~B.1]{lu2025simplifying}. Setting $s=0$ in \eqref{eq:trigflow_ddim} recovers the CM parameterization \eqref{eq:cm_param}; setting $s=\tk$ and adding segment conditioning recovers \eqref{eq:msscd_cm}.

\subsection{ADN normalization}
\label{app:adn}

We restate EDM2's \texttt{normalize} function and derive its properties.

\textbf{Function:}
\begin{align}
  \norm(x,\,\dim) &= x \,/\, (N(x) + \varepsilon), \nonumber \\
  N(x) &:= \|x\|_{\ell_2,\,\dim} \cdot \sqrt{\frac{|\text{norm}(x)|}{|x|}}, \quad \varepsilon = 10^{-4},
\end{align}
where $\|\cdot\|_{\ell_2,\dim}$ denotes the $\ell_2$ norm along the specified axes (with \texttt{keepdim=True}), $|\text{norm}(x)|$ is the number of elements in the norm output tensor, and $|x|$ is the total number of elements in $x$.

\textbf{Applied along \texttt{dim=1} to shape $[B,C,H,W]$:} The norm output has shape $[B, 1, H, W]$, so $|\text{norm}| = B{\cdot}H{\cdot}W$ and $|x| = B{\cdot}C{\cdot}H{\cdot}W$, giving $\sqrt{|\text{norm}|/|x|} = 1/\sqrt{C}$. Therefore $N(x) = \|v\|_2 / \sqrt{C} = \mathrm{RMS}(v)$ (where $v = x_{b,:,h,w}$ is the channel vector at spatial location $(b,h,w)$).

Dividing by the Euclidean norm alone gives an RMS of $1/\sqrt{C}$, rather than approximately one. sCM cites \citet{karras2017progressive} for $\pnorm$ but does not fix $\varepsilon$ or its placement. We use EDM2's \texttt{normalize} ($\varepsilon=10^{-4}$, added outside the norm) for the EDM2 backbone and the original PGGAN form ($\varepsilon=10^{-8}$, inside the square root) for DDPM++.

\textbf{ADN in the EDM2 residual block:}
\begin{align}
  [s_{\text{pre}}, b_{\text{pre}}] &= \mathrm{split}(\texttt{emb\_linear}(\mathrm{emb})), \\
  s &= \norm(s_{\text{pre}}, \mathrm{dim}=1), \quad b = \norm(b_{\text{pre}}, \mathrm{dim}=1), \\
  y &= \norm(y, \mathrm{dim}=1) \odot s[\ldots] + b[\ldots],
\end{align}
where $[\ldots]$ denotes broadcasting over $H,W$. DDPM++ uses GroupNorm for the features and the PGGAN pixel normalization defined above for scale and shift.

\section{Dropout in Multistep Consistency Models}
\label{app:sync_dropout}

Discrete MSCD uses two student forward passes per step. We synchronize their dropout masks by saving and restoring the random state, following \citet{song2024improved}. This prevents dropout randomness from contributing to the consistency residual. In our MSCD reproductions on EDM \citep{karras2022elucidating} and EDM2 \citep{karras2024analyzing}, students without dropout overfit within the training budget.

Our single-step sCD runs use zero dropout, following \citet[Appendix~G.2]{lu2025simplifying}. Continuous MS-sCD computes the prediction and its time derivative in one JVP forward pass, so they share a dropout mask without explicit synchronization. For MS-sCD, no dropout led to overfitting and an eight-step FID near $2.3$; a rate of $0.2$ slowed training and showed signs of instability. We use $0.1$ at resolutions $\le 16$ for all reported MS-sCD results.

\section{Noise Sampling for the Multistep Students}
\label{app:noise_policy}

\citet{heek2024multistep} split diffusion time into equal segments and sample the training noise uniformly over a discrete, cosine-parameterized step grid that anneals from $64$ to $1280$ steps. We instead use framework-specific noise schedules. For the discrete MSCD on EDM2-S we use the teacher's EDM (Karras) $\sigma$ grid, annealing the step count from $64$ to $1280$. For MS-sCD we keep the segment boundaries uniform in TrigFlow time but sample $\sigma$ from a log-normal proposal ($P_{\text{mean}}=-1.0$, $P_{\text{std}}=1.6$; teacher: $P_{\text{mean}}=-0.8$) and set $t=\arctan(\sigma/\sd)$, instead of spreading samples evenly across segments. This global proposal worked better in our runs. Spreading samples evenly across segments puts too much weight on noise levels the teacher rarely saw in training.

For the two-step ($M{=}2$) discrete MSCD on EDM2 we set the intermediate segment boundary of the student grid to $\sigma=1.1$, giving $[\sigma_{\max}, 1.1, 0]$, since the default Karras schedule places it near $\sigma_{\min}=0.002$ and collapses the first segment. The $M{=}4$ runs keep the default Karras grid.

\section{Matched Segment-Boundary Ablation for MSCD vs.\ MS-sCD}
\label{app:matched_grid}

We retrain MSCD using the same segment boundaries as MS-sCD, converting TrigFlow time to EDM noise levels through the correspondence of \citet{lu2025simplifying}:
\begin{equation}
  t = \arctan(\sigma/\sd), \qquad \sigma = \sd\tan(t), \qquad x_t = \cos(t)\,x_\sigma.
\end{equation}
This places the boundaries at the same noise levels in both methods; all other MSCD hyperparameters are unchanged. With $\sd=0.5$, $\sigma_{\max}=80$, and $\tmax=\arctan(160)\approx 1.5645$, the intermediate segment boundaries are:
\begin{center}
  \begin{tabular}{ll}
    \toprule
    $M$ & Intermediate boundaries (descending $\sigma$) \\
    \midrule
    2 & $0.4969$ \\
    4 & $1.1913,\ 0.4969,\ 0.2062$ \\
    8 & $2.4437,\ 1.1913,\ 0.7420,\ 0.4969,\ 0.3324,\ 0.2062,\ 0.0991$ \\
    \bottomrule
  \end{tabular}
\end{center}

At $M{=}2$, under a similar training budget, the matched grid reaches best FID $3.64$ versus $3.44$ for the original $\sigma$-space grid in Table~\ref{tab:msscd}. At $M{=}4$, the matched grid again trails ($3.20$ versus $2.97$ at comparable budget; original full-budget FID $2.32$). At $M{=}8$, under a similar training budget, the matched grid reaches $1.99$ versus $1.79$ for the original grid. Using the same segment boundaries did not improve MSCD in these runs. We therefore retain the original schedules for the comparison in Table~\ref{tab:msscd}.

\section{Iterative Single-Step sCD Baseline}
\label{app:iter_scd}

MS-sCD trains a segment-conditioned student and samples with one network evaluation per segment. A useful control for assessing the benefits of MS-sCD is to keep the one-step sCD checkpoint fixed and apply Song-style iterative sampling \citep[Algorithm~1]{song2023consistency}, adapted to TrigFlow \citep{lu2025simplifying}: first predict $\hat x_0=f_\theta(x_{\tmax},\tmax)$, then draw fresh $z\sim\mathcal{N}(0,\sd^2I)$, form $x_{t_{\mathrm{mid}}}=\cos(t_{\mathrm{mid}})\hat x_0+\sin(t_{\mathrm{mid}})z$, and update $\hat x_0=f_\theta(x_{t_{\mathrm{mid}}},t_{\mathrm{mid}})$. Repeat the re-noising and prediction pair until a total of $M$ network evaluations have been used. We keep $t_{\mathrm{mid}}=1.1$ fixed across repetitions, rather than optimizing a decreasing sequence of intermediate times. This uses the same sampler as our NFE${=}2$ evaluation; it is \emph{not} the deterministic MS-sCD segment chain of Algorithm~\ref{alg:msscd_sample}. Table~\ref{tab:iter_scd} shows that additional iterations improve sCD's FID, but MS-sCD performs better at the same NFE.

\begin{table}[h]
  \centering
  \caption{Iterative single-step sCD vs.\ MS-sCD on ImageNet-64 (EDM2-S, class-conditional, no CFG, 50k samples). The sCD results at NFE${=}1$/$2$ are from Table~\ref{tab:students}; MS-sCD results are from Table~\ref{tab:msscd}.}
  \label{tab:iter_scd}
  \begin{tabular}{lcc}
    \toprule
    Method & NFE & FID \\
    \midrule
    sCD iterated & 1 & 3.59 \\
    sCD iterated & 2 & 2.66 \\
    sCD iterated & 4 & 2.48 \\
    sCD iterated & 8 & 2.32 \\
    \midrule
    MS-sCD & 2 & 2.53 \\
    MS-sCD & 4 & 2.26 \\
    MS-sCD & 8 & 2.08 \\
    \bottomrule
  \end{tabular}
\end{table}

\section{MS-sCD on CIFAR-10 and Teacher NFE}
\label{app:msscd_cifar}

Table~\ref{tab:msscd_cifar} extends MS-sCD to unconditional CIFAR-10. Increasing $M$ improves FID, but $M{=}8$ ($2.20$) does not surpass our two-step sCT ($2.09$). A useful reference point is the teacher ODE itself: the same TrigFlow checkpoint that reaches FID $2.08$ at the standard 35-NFE DPM-Solver-2S setting \citep{lu2025simplifying,karras2022elucidating} evaluates to FID $2.22$ at 199 NFE. For an exact PF-ODE, more steps should reduce numerical integration error \citep{karras2022elucidating,lu2022dpm}; FID, however, measures samples from the \emph{learned} vector field, so its dependence on NFE need not be monotone when model error or solver/schedule mismatch dominates. EDM likewise selects NFE by near-minimum observed FID rather than assuming a monotone curve \citep{karras2022elucidating}. MS-sCD's CIFAR-10 $M{=}8$ result matches that denser 199-NFE teacher evaluation within seed variation ($2.20$ vs.\ $2.22$), suggesting that learned-ODE quality may influence this result under finer integration.

\section{Experimental Details}
\label{app:experimental_details}

\subsection{Teacher training}
\label{sec:teacher:training}

Both teachers are trained with the TrigFlow diffusion loss \eqref{eq:teacher_loss} using log-normal noise-level sampling $t = \arctan(e^\tau/\sd)$, $\tau \sim \mathcal{N}(P_{\text{mean}}, P_{\text{std}}^2)$, $\sd = 0.5$.

\textbf{ImageNet-64 (EDM2-S backbone).} $P_{\text{mean}} = -0.8$, $P_{\text{std}} = 1.6$. Optimizer: Adam, $\beta_1=0.9$, $\beta_2=0.99$, $\varepsilon=10^{-11}$, reference learning rate $\eta_{\text{ref}} = 0.01$ (EDM2-style auto-scaling via reference batch count 35{,}000). Batch size 2048 on 32 H100 GPUs. We use the post-hoc power-function EMA framework of EDM2 (phEMA). The training run records phEMA snapshots, and at evaluation we synthesize the EMA at a chosen $\sigma_{\text{rel}}$ \citep{karras2024analyzing}. The sCM paper's appendix uses $\sigma_{\text{rel}}=0.075$. In our run we swept $\sigma_{\text{rel}}$ and found the optimal value to be $\sigma_{\text{rel}}{=}0.115$, which we use for the reported FID. The remaining training settings match the sCM appendix.

\textbf{CIFAR-10 (DDPM++ backbone, unconditional).} We follow the teacher settings in Appendix~G.1 of \citet{lu2025simplifying}, which themselves follow the EDM~2022 CIFAR-10 settings \citep{karras2022elucidating}: $P_{\text{mean}} = -1.2$, $P_{\text{std}} = 1.2$, dropout 0.13, batch size 512, learning rate $10^{-3}$, Adam $\beta_1=0.9, \beta_2=0.999, \varepsilon=10^{-8}$, no learning-rate decay, 400k training iterations. Backbone: Score~SDE DDPM++ \citep{song2021score} with channel multiplier $[2,2,2]$, four residual blocks per resolution, and attention at $16{\times}16$.

For CIFAR-10 we use EDM's augmentation pipeline (horizontal flip, isotropic scale, rotation, anisotropic scale, fractional translation, all at $p=0.12$) with the augmentation labels passed to the network. ImageNet-64 uses no augmentation.

\begin{table}[h]
  \centering
  \caption{Hyperparameters for all experiments.}
  \label{tab:hparams}
  \begin{tabular}{lcccccc}
    \toprule
    & Teacher & sCD & sCT & MS-sCD & MSCD-EDM & MM \\
    \midrule
    Framework & TrigFlow & TrigFlow & TrigFlow & TrigFlow & EDM & EDM \\
    Model & EDM2-S & EDM2-S & EDM2-S & EDM2-S & EDM2-S & EDM-S \\
    Batch size & 2048 & 2048 & 2048 & 2048 & 2048 & 2048 \\
    GPUs & 32 & 32 & 32 & 32 & 32 & 16 \\
    $P_{\text{mean}}$ & $-0.8$ & $-1.0$ & $-1.0$ & $-1.0$ & $-0.8$ & --- \\
    $P_{\text{std}}$ & $1.6$ & $1.6$ & $1.6$ & $1.6$ & $1.6$ & --- \\
    $\sd$ & $0.5$ & $0.5$ & $0.5$ & $0.5$ & $0.5$ & $0.5$ \\
    Dropout & 0 & 0 & 0.45 & 0.10 & 0.10 & 0 \\
    LR ref & $0.01$ & $1\times10^{-4}$ & $1\times10^{-4}$ & $5\times10^{-4}$ & $5\times10^{-4}$ & $2\times10^{-6}$ \\
    Adam $\beta$ & (0.9, 0.99) & (0.9, 0.99) & (0.9, 0.99) & (0.9, 0.99) & (0.9, 0.99) & (0.0, 0.99) \\
    Adam $\varepsilon$ & $10^{-11}$ & $10^{-11}$ & $10^{-11}$ & $10^{-11}$ & $10^{-8}$ & $10^{-12}$ \\
    phEMA std & 0.075 & 0.075 & 0.075 & --- & --- & --- \\
    Eval.\ weights & phEMA & phEMA & phEMA & both & EMA & live \\
    Tangent $c$ & --- & 0.1 & 0.1 & 0.1 & --- & --- \\
    Warmup $H$ & --- & 10k & 10k & 10k & --- & --- \\
    Params & 293.9M & 293.9M & 293.9M & 293.9M & 280.2M & 295.9M \\
    \bottomrule
  \end{tabular}
\end{table}

\textbf{Weight normalization during JVP calls.} EDM2's magnitude-preserving convolutions normalize the stored weights by their RMS before each training forward pass and apply functional normalization when computing the output. We disable the in-place weight update during JVP calls. The JVP retains the functional weight normalization, but the stored weight norms can drift during student training. We have not observed instability attributable to this drift.

\textbf{Live weights versus phEMA.} For the teacher and single-step students, we sweep the post-hoc EMA (phEMA) width $\sigma_{\text{rel}}$. For MS-sCD, we also evaluate live (non-EMA) weights saved every $4\,\mathrm{Mi}$ (${\approx}4.2$M) training images. In each run, we report the checkpoint and weight choice with the lowest validation FID.

\textbf{FID evaluation.} We evaluate Fr\'echet Inception Distance (FID) \citep{heusel2017fid} using 50,000 samples with the standard ADM evaluation protocol \citep{dhariwal2021diffusion}, compared against the ILSVRC 2012 training set statistics. Generation is class-conditional for ImageNet-64, with no classifier-free guidance. Following \citet{karras2024analyzing}, we compute FID over a sweep of generator seeds and report the best in the main tables. Table~\ref{tab:seed_stats} additionally reports mean $\pm$ sample standard deviation across seeds. Table~\ref{tab:prdc} complements FID with precision/recall \citep{kynkaanniemi2019improved} and density/coverage \citep{naeem2020reliable} on the same 50k-sample sets used for those best-seed FIDs. Density and coverage follow the FID ordering (teacher ${>}{\,}$MS-sCD $M{=}8$ ${>}{\,}$ $M{=}4$ ${\approx}$ $M{=}2$ ${>}{\,}$sCD-2 ${>}{\,}$sCD-1 ${\gg}$\,sCT).

\begin{table}[h]
  \centering
  \caption{ImageNet-64 seed FID mean $\pm$ std (EDM2-S, class-conditional, no CFG, 50k samples). Best matches the main-text FID.}
  \label{tab:seed_stats}
  \begin{tabular}{lccc}
    \toprule
    Model & NFE & Best & Mean $\pm$ std \\
    \midrule
    TrigFlow teacher & 63 & 1.83 & $1.8334 \pm 0.0005$ \\
    sCD & 1 & 3.59 & $3.6423 \pm 0.0330$ \\
    sCD & 2 & 2.66 & $2.6856 \pm 0.0153$ \\
    sCT & 1 & 4.34 & $4.4403 \pm 0.0557$ \\
    sCT & 2 & 3.98 & $4.0093 \pm 0.0198$ \\
    MS-sCD ($M{=}2$) & 2 & 2.53 & $2.5761 \pm 0.0231$ \\
    MS-sCD ($M{=}4$) & 4 & 2.26 & $2.2853 \pm 0.0199$ \\
    MS-sCD ($M{=}8$) & 8 & 2.08 & $2.1227 \pm 0.0243$ \\
    \bottomrule
  \end{tabular}
\end{table}

\begin{table}[h]
  \centering
  \caption{ImageNet-64 precision (P), recall (R), density (D), and coverage (C) on the best-seed 50k-sample sets (EDM2-S, class-conditional, no CFG), using unnormalized 2048-d Inception-2015 features and $k{=}5$ \citep{naeem2020reliable}.}
  \label{tab:prdc}
  \begin{tabular}{lcccccc}
    \toprule
    Model & NFE & FID & P & R & D & C \\
    \midrule
    TrigFlow teacher & 63 & 1.83 & 0.802 & 0.708 & 1.165 & 0.968 \\
    sCD & 1 & 3.59 & 0.776 & 0.703 & 1.062 & 0.950 \\
    sCD & 2 & 2.66 & 0.780 & 0.708 & 1.104 & 0.960 \\
    sCT & 1 & 4.34 & 0.722 & 0.737 & 0.878 & 0.900 \\
    sCT & 2 & 3.98 & 0.719 & 0.747 & 0.890 & 0.904 \\
    MS-sCD ($M{=}2$) & 2 & 2.53 & 0.785 & 0.702 & 1.121 & 0.960 \\
    MS-sCD ($M{=}4$) & 4 & 2.26 & 0.789 & 0.705 & 1.127 & 0.964 \\
    MS-sCD ($M{=}8$) & 8 & 2.08 & 0.788 & 0.707 & 1.147 & 0.966 \\
    \bottomrule
  \end{tabular}
\end{table}

\section{Uncurated Samples on ImageNet-64}
\label{app:samples}

Each panel below shows 30 uncurated class-conditional ImageNet-$64{\times}64$ samples (arranged as a $6 \times 5$ grid) from one of the four models discussed in the paper.

\begin{figure}[h]
  \centering
  \begin{subfigure}[t]{0.48\linewidth}
    \centering
    \includegraphics[width=\linewidth]{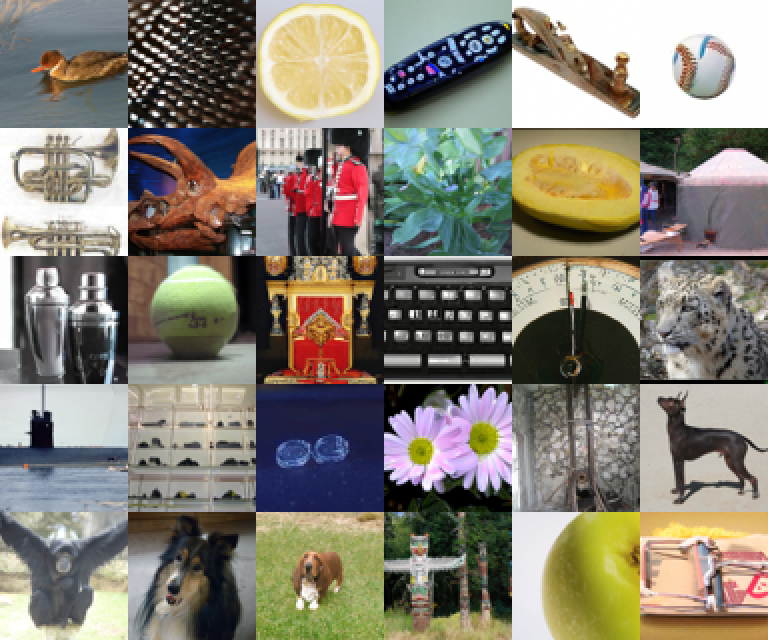}
    \caption{TrigFlow teacher (EDM2-S backbone), 63-NFE DPM-Solver-2S samples.}
    \label{fig:samples_teacher}
  \end{subfigure}\hfill
  \begin{subfigure}[t]{0.48\linewidth}
    \centering
    \includegraphics[width=\linewidth]{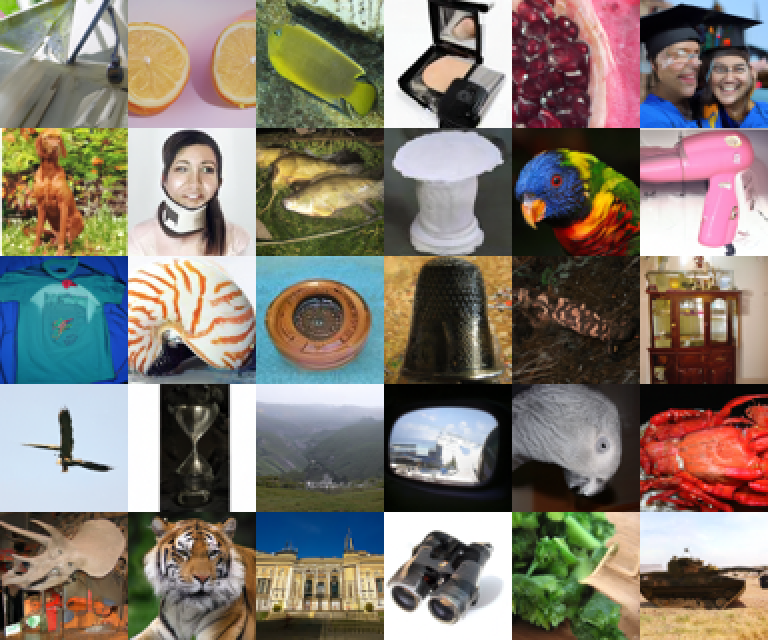}
    \caption{sCD student (TrigFlow, EDM2-S, ours), 2-NFE samples with intermediate time $t_{\text{mid}}=1.1$ (the iterative sampler in Appendix~\ref{app:iter_scd}).}
    \label{fig:samples_scd}
  \end{subfigure}
  \caption{Teacher and sCD student samples.}
  \label{fig:samples_pair1}
\end{figure}

\begin{figure}[h]
  \centering
  \begin{subfigure}[t]{0.48\linewidth}
    \centering
    \includegraphics[width=\linewidth]{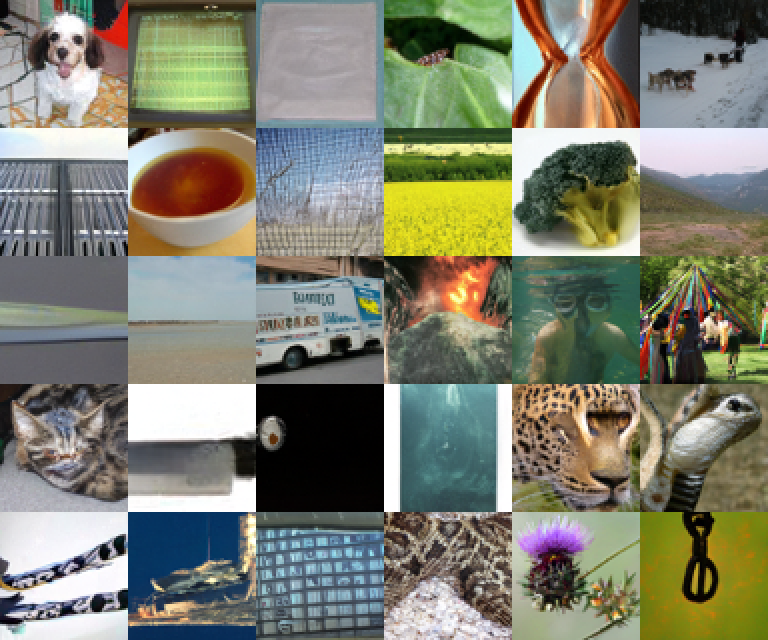}
    \caption{sCT student (TrigFlow, EDM2-S, ours), 2-NFE samples with intermediate time $t_{\text{mid}}=1.1$ (the iterative sampler in Appendix~\ref{app:iter_scd}).}
    \label{fig:samples_sct}
  \end{subfigure}\hfill
  \begin{subfigure}[t]{0.48\linewidth}
    \centering
    \includegraphics[width=\linewidth]{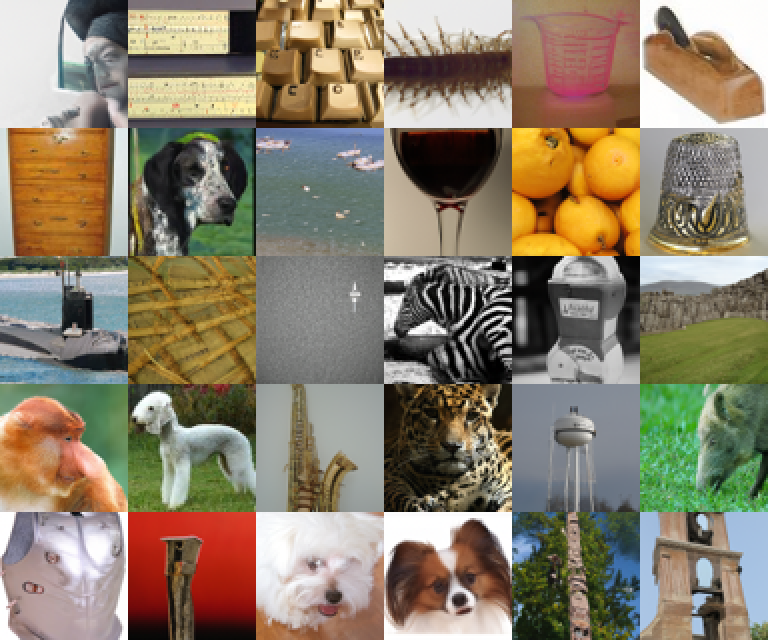}
    \caption{MSCD on EDM2-S (ours), $M=8$ (8-NFE) samples.}
    \label{fig:samples_mscd_m8}
  \end{subfigure}
  \caption{sCT student and 8-step MSCD samples.}
  \label{fig:samples_pair2}
\end{figure}

\end{document}